# LittleYOLO-SPP: A Delicate Real-Time Vehicle Detection Algorithm

[1]Sri Jamiya S  [2]Esther Rani P

**Abstract:** Vehicle detection in real-time is a challenging and important task. The existing real-time vehicle detection lacks accuracy and speed. Real-time systems must detect and locate vehicles during criminal activities like theft of vehicle and road traffic violations with high accuracy. Detection of vehicles in complex scenes with occlusion is also extremely difficult. In this study, a lightweight model of deep neural network LittleYOLO-SPP based on the YOLOv3-tiny network is proposed to detect vehicles effectively in real-time. The YOLOv3-tiny object detection network is improved by modifying its feature extraction network to increase the speed and accuracy of vehicle detection. The proposed network incorporated Spatial pyramid pooling into the network, which consists of different scales of pooling layers for concatenation of features to enhance network learning capability. The Mean square error (MSE) and Generalized IoU (GIoU) loss function for bounding box regression is used to increase the performance of the network. The network training includes vehicle-based classes from PASCAL VOC 2007,2012 and MS COCO 2014 datasets such as car, bus, and truck. LittleYOLO-SPP network detects the vehicle in real-time with high accuracy regardless of video frame and weather conditions. The improved network achieves a higher mAP of 77.44% on PASCAL VOC and 52.95% mAP on MS COCO datasets.

[1]Sri Jamiya S, Research Scholar, Electronics and Communication Engineering, Vel Tech Rangarajan Dr. Sagunthala R &D Institute of Science and Technology, Avadi, Chennai. India.
Email: vtd655@veltech.edu.in

[2]Esther Rani P, Professor, Electronics and Communication Engineering, Vel Tech Rangarajan Dr. Sagunthala R &D Institute of Science and Technology, Avadi, Chennai. India.
Email: drpestherrani@veltech.edu.in

## 1. Introduction

Vehicle detection is a significant part of the problem in networks such as the development of an autonomous driving system, real-time surveillance systems, and Intelligent transportation systems. Autonomous driving is an emerging high-tech innovation that needs to locate mainly vehicles rather than other objects. With the rapid increase of vehicles in the urban and city places, there are numerous incidents regarding traffic violations, vehicle accidents, and vehicle thefts which should be monitored using CCTV cameras. For Traffic surveillance systems, the detector must be quick to detect vehicles in real-time. There are various developments in the field of Traffic Management systems and Video surveillance technologies. The detection of anomalies in traffic surveillances such as traffic congestions, parking violations, and rash driving on the roads is one of the finest works in vehicle detection [1].

Machine learning models for object detection networks are less attractive due to their complex nature and advancements of deep learning neural networks for detection. Deep learning networks training speed is significantly less in CPU computation, but due to the emerging technologies like GPU and TPU, the training time is much reduced. Object detection algorithms are a fundamental approach to vehicle detections. The three main kinds of vehicle detection models are motion-based models, feature-based models, and convolutional neural network-based models.

Motion-based models include background subtraction, optical flow, and frame subtraction. The background subtraction network of vehicle detection is based on GMM [2]. The GMM differentiates the foreground and background to detects the vehicles, but it is not ideal for detecting stationary vehicles. In optical flow method pixels of feature vectors are located by the model, then tracking of these vectors pixels are done. It is time-consuming and has complexity issues [3]. Frame subtraction techniques are done by calculating differences between the two or more frames to detect the vehicles in motion, but it is not the right choice for swift and passive movements.

Feature-based models are HOG, Haar, and SURF. The Histogram of oriented gradients feature has a rich descriptive nature which groups the pixels according to the gradients. The orientation of the objects pixel is extracted by the HOG, which is used for detection. The region of interest is pointed using the Haar features [4]. The detection of features like recognition of vehicle parts, reconstruction of 3D, recognition of objects, and classification is processed by SURF. It is derived from the descriptor of (Scale Invariant Feature Transform) SIFT. SURF is more efficient, robust, and has the best performance than SIFT. SURF is highly effective in object labelling [5].

Earlier deformable part-based model (DPM) [6] performed better in detection using features of HOG and classifiers such as Adaboost and Support vector machines. Then after the introduction of two-step approached detectors and one stage detectors, these models and features are not used anymore. Convolutional neural network approaches such as R-CNN uses selective search to generate the region of proposals [7]. Then some advancements are made, and Fast R-CNN, Faster RCNN, SPP-net models are emerged achieving state-of-the-art results [8-10]. Mask RCNN is an extension to Faster RCNN designed to address issues in locating pixel-wise object instance segmentation and has a low detection rate [11]. Overfeat network is fast in detection but lacks in accuracy than RCNN. Tracking of detected objects is vital to analyze and predict the motion of vehicles. There are CNN based and RNN based trackers [12]. CNN trackers are excellent in tracking due to its feature extraction and classification techniques. Some examples of CNN trackers are VGGNet, DeepSRDCF, and Faster-RCNN.

The main issues related to the detection of vehicles are lighting conditions and the quality of the video feed. During day vehicles can be easily detected by visible factors such as vehicle front, rear, edges, and corners. But at night time these features are not visual, hence the trained detector could not perform well. A robust detector should be built with less complexity to rectify these kinds of issues. Moreover, some vehicles which are very far could also have not been easily detected by the detector due to its small size.

The primary purpose of this paper is to design a lightweight model of robust vehicle detection. Existing works are mainly focused on detecting all objects rather than vehicles. Vehicle detection is a special case. If the object detection networks are used for detecting vehicles, it will not be in an effective manner. Vehicle detection requires low-level feature maps, a small aspect ratio of images, and it should be fast enough to predict in real-time with high accuracy. With all the necessary details for designing the vehicle detection network, a suitable network is selected as a base for building this vehicle detector network model. First, a literature review is conducted to reveal related investigations. Then the proposed method is explained in the next section. The subsections disclose the performance of the proposed approach, and comparisons are also conducted. Finally, this study concludes with the discussions and conclusion.

## 2. Related works

Vehicle detections mostly rely on Object detection algorithms. In computer vision, object detection is one of the popular topics. In recent years many advancements have been made in object detection algorithms. Object detection networks now can predict objects with the best accuracy rate and higher frames per second in speed. The two kinds of approaches in the object detection algorithms are two-stage and one-stage approaches [13]. The old-fashioned detection networks are two-stage approaches; they are highly accurate in prediction but lacks in performance and speed. The networks which fall under this type are R-CNN, Fast R-CNN, Faster RCNN, and SPP-net [7-10]. These networks first create a group of object proposals and then predicts the object regions and their corresponding labels. The one-stage detectors made numerous achievements in the popular PASCAL VOC and MS COCO datasets [14,15]. The limitations of these two-stage approaches are speed due to its localization in the first stage and detection in the second stage.

The best models in the object detection networks are one stage approaches. The performance of these models is phenomenal. They are speedy and predict objects with high efficiency. The networks which come under this category are YOLO, YOLOv2, YOLOv3, and SSD [16-19]. In YOLO the network is made up of single feed-forward convolutional layers that can detect objects classes and locations directly in one flow. YOLOv2 consists of batch normalization, the classifier with high resolution, bounding boxes were predicted by anchor boxes, and a custom backbone network called Darknet19. In YOLOv3 the network prediction is made by three different scales which makes the network to be more effective than its earlier versions. YOLOv3 uses Darknet53 as a Feature extraction network and logistic regression instead of the softmax layer. These networks speed is high but can't able to perform better in the night and complex environmental conditions such as in fog and rain.

The recent advancements in the object detection deep neural networks are RetinaNet [20], Cascade R-CNN, CenterNet, RefineDet, and YOLOv4 [21-24]. These networks record challenging accuracy in popular PASCAL VOC and MS COCO datasets. Cascade R-CNN [21] deals with the issues in the network thresholding and inference time mismatch. It is a multi-stage detection network with each output from the detector placed as a training input of other detectors. It reduces overfitting and rectifies mismatch between detectors and hypothesis. In keypoints based object detection, there is colossal mismatching between bounding boxes occurs due to inadequate processing. CenterNet is formed from the keypoint point-based

CornerNet. It improves accuracy by detecting objects as triplets instead of key points [22]. It also enhances efficiency by using two pooling methods cascade corner pooling and centre pooling. Two-stage approaches are capable of delivering high accuracy, whereas the one-stage strategies provide high efficiency, to balance both Refine Detection, single-shot based networks are formed [23]. The two different modules in the network structure are anchor refinement module and a detection module. The detector in the network first refines the anchors based on negative anchors and adjusts the location of anchor depends on sizes for predicting using improved regression.YOLOv4 [24] paper explains various aspects and methods to improve object detection network's accuracy and efficiency. It uses multiple new features such as Weighted-Residual-Connections, Cross mini-Batch Normalization, DropBlock regularization, and CIoU loss to enhance the networks.

The few works related to YOLOv3-tiny includes YOLO-LITE, YOLO Nano, Slim YOLO, and Spiking-YOLO[25-28]. YOLO-LITE a very lightweight seven-layered network achieves higher accuracy than SSD Mobilenetv1. It is based on the YOLOv2 algorithm. It is designed in a way to run smoother on Non-GPU based computers.YOLO Nano a compact micro model of object detection based on YOLO and SSD used for embedded applications. Nano's architecture size is 4MB which is smaller than YOLOv2-tiny and YOLOv3-tiny networks and also achieves higher mAP than these two networks. SlimYOLO performs pruning on channels of convolutional layers. It is trained to detect UAVs with fewer parameters than YOLOv3-tiny. It achieves remarkable performance with SPP on the drones dataset. These methods are typically designed to detect objects which do not depend on which input size, feature extractor to use for quick and accurate detection. In this paper, we focused on these factors and achieved better trade-off between speed and accuracy.

Many systems adopt Spiking neural networks due to its less power utilization and event-driven functions. But, the SNN networks are hard to train and only useful in object classification. Although these networks made significant improvements, it is not up to the mark to be used for autonomous driving systems. Still, further enhancements and less complex design of networks are necessary. For vehicle detection, the network must be simple, and its computational cost should not be high. The vehicle detection networks are a subcategory of object detection algorithms. The detector of object detection networks is usually trained with all classes in a dataset. The PASCAL VOC dataset contains 20 classes, and the MS COCO dataset contains 80 classes, for object detection all classes are included for training. In vehicle detection, the networks are trained with vehicle classes such as bicycle, car, motorbike, aeroplane, bus, train, truck, and boat. In the Traffic management systems and Intelligent transportation systems classes such as car, bus, and truck are used for training the vehicle detection network.

## 3. Proposed work

The YOLOv3-tiny network is a miniature version of the YOLOv3 object detection network. YOLOv3 and YOLOv3-tiny both are designed for the task of detecting multiple classes rather than one particular class. These networks are trained on the MS COCO dataset and achieved better results. The proposed system consists of an improved YOLOv3-tiny network, Spatial pyramid pooling layers [29] with GIoU [30] is used for regression loss and also as a metric for evaluation. The k-means++ algorithm is used for selecting appropriate bounding boxes for the predicted objects. The network specifically designed for vehicle detection trained only with vehicle classes from datasets. The architecture of the LittleYOLO-SPP is drawn in Fig.1.

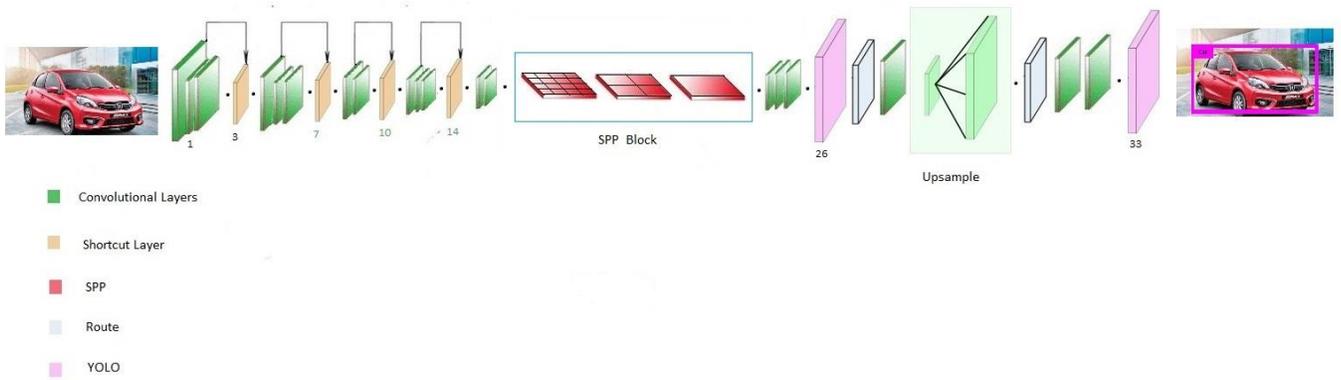

**Fig. 1.** The architecture of LittleYOLO-SPP network model

*3.1 Network Structure of LittleYOLO-SPP*

The network YOLO is introduced by Redmon et al. [16] which processes the image in one flow to detect objects quickly. To build one of the best object detection model feature extraction network is very important. If a network extracts more number of features, then the network is capable of detecting objects with a high accuracy rate. YOLO uses GoogleNet [31] as a feature extraction network. Darknet 19 and Darknet53 are convolutional neural networks used as a Feature extraction network in YOLOv2 and YOLOv3 [17, 18]. The network of YOLOv3-tiny is fast but not so accurate in detections. Hence, some enhancements are necessary for building a robust vehicle detection network.

*3.2 Feature extraction layers*

The original YOLOv3-tiny network has ten convolutional layers and six max-pooling layers as a feature extractor. These convolutional layers follow the YOLOv3-tiny detection layers. The detailed information about the network is shown in Table 1 [18]. Though this network performs better in object detection, its processing speed and computational complexity are high. For the vehicle detection model, the layers of the YOLOv3-tiny network should be modified to build a reliable vehicle detector.

LittleYOLO-SPP consists of 13 convolutional layers with shortcut connections in between them. The feature maps are downsampled just after each shortcut layer. The shortcut layer is nothing but just a residual connection in the network. The max-pooling layers are removed from the YOLOv3-tiny network's convolutional layers and then incorporated the Spatial pyramid pooling network after the feature extraction layers. The network is designed to extract rich features for detection at the same time the process will not be slow. The SPP minimizes the computational complexity. It forms the feature maps from the whole image in a single pass. The features of convolutional layers are not processed again and again while using SPP, hence the network speed increases. The entire architecture has a depth of 33 layers; all the details of the proposed network are listed in Table 2.

*3.3 Anchor boxes using K-means++ Clustering Algorithm*

The YOLOv3-tiny uses k-means clustering for selecting anchor boxes. The k-means algorithm cluster objects based on its attributes. It is simple and easy to understand, which is widely used by many network models. The k-means is a distance-based clustering algorithm. Every cluster is defined by centre of the cluster and the meeting point of the centroid of those clusters. It is initiated by predicting a seed point nearest to the cluster for each object and compute seed points as the centroid of the current mean point of the cluster. It has many disadvantages, k number of clusters must be specified in advance, and it will not perform well in noisy images. Therefore, the anchors are calculated from the k-means++ algorithm, which is better than the k-means algorithm. In k-means++, the centroids are initiated randomly, then the distance is calculated between initial datapoint with all other datapoints. Then select all other datapoints one by one and calculate the distances between them like the initial step.

The k-means++ algorithm is used on the PASCAL VOC and MS COCO datasets to calculate anchor boxes with 6 clusters. The calculated six anchors for image size 640×640 are (25×23), (69×59), (123×141), (290×159), (275×339), (526×450). The anchor boxes size for input size 416 × 416 are (16×15), (42×40), (95×73), (115×165), (256×168), (329×314). The first three anchors are intended for detecting smaller vehicles on the final YOLO layer, and the next three anchors are used for larger size vehicles on the first YOLO detection layer.

**Table 1 YOLOv3-tiny Network**

| Type | Filters | Size/Stride | Output |
|---|---|---|---|
| Convolution | 16 | 3 x 3 | 416 x 416 |
| Max | | 2 x 2/ 2 | 208 x 208 |
| Convolution | 32 | 3 x 3 | 208 x 208 |
| Max | | 2 x 2/ 2 | 104 x 104 |
| Convolution | 64 | 3 x 3 | 104 x 104 |
| Max | | 2 x 2/ 2 | 52 x 52 |
| Convolution | 128 | 3 x 3 | 52 x 52 |
| Max | | 2 x 2/ 2 | 26 x 26 |
| Convolution | 256 | 3 x 3 | 26 x 26 |
| Max | | 2 x 2/ 2 | 13 x 13 |
| Convolution | 512 | 3 x 3 | 13 x 13 |
| Max | | 2 x 2 | 13 x 13 |
| Convolution | 1024 | 3 x 3 | 13 x 13 |
| Convolution | 256 | 1 x 1 | 13 x 13 |
| Convolution | 512 | 3 x 3 | 13 x 13 |
| Convolution | 255 | 1 x 1 | 13 x 13 |
| YOLO | | | |
| Route | | | 13 x 13 |
| Convolution | 128 | 1 x 1 | 13 x 13 |
| Upsample | | 2x | 26 x 26 |
| Route | | | 26 x 26 |
| Convolution | 256 | 3 x 3 | 26 x 26 |
| Convolution | 255 | 1 x 1 | 26 x 26 |
| YOLO | | | |

*3.4 LittleYOLO-SPPDetection Network*

In LittleYOLO-SPP, the prediction is performed by the YOLO detection networks. The network is made up of a few convolutional layers with an upsampling factor of two. The bounding boxes of vehicles are predicted using Anchor boxes defined by the k-means++ algorithm. In the YOLO layer, bounding box, objectness score, anchors, and class predictions are processed. For each bounding box, the network predicts four coordinates. The kernel size for the convolutional layer before the YOLO layer is calculated by 1 x 1 x (Bb x (5 + num of Classes)). Bb is bounding boxes is 3, (4 + 1 = 5) is derived from 4 bounding box offsets and 1 objectness score. The total number of classes used here are two, car and bus from PASCAL VOC and three classes car, bus, and truck from MS COCO. Therefore, the kernel size of the layer is 1 x 1 x 21 and 1 x 1 x 24, respectively. The feature map is upsampled by 2x, which is selected from two layers above. From SPP, fine-grained features are extracted from the layers. Then the final feature maps are used to detect the vehicles by YOLO layers.

**Table 2 LittleYOLO-SPPNetwork**

| Type | Filters | Size/Stride | Output |
|---|---|---|---|
| Convolution | 16 | 3 x 3 | 416 x 416 |
| Convolution | 32 | 3 x 3/ 2 | 208 x 208 |
| Convolution | 64 | 3 x 3 | 208 x 208 |
| Shortcut Layer | | | |
| Convolution | 32 | 3 x 3/ 2 | 104 x 104 |
| Convolution | 64 | 3 x 3 | 104 x 104 |
| Convolution | 128 | 3 x 3 | 104 x 104 |
| Shortcut Layer | | | |
| Convolution | 256 | 3 x 3/ 2 | 52 x 52 |
| Convolution | 128 | 3 x 3 | 52 x 52 |
| Shortcut Layer | | | |
| Convolution | 256 | 3 x 3/ 2 | 26 x 26 |
| Convolution | 512 | 3 x 3 | 26 x 26 |
| Convolution | 256 | 3 x 3 | 26 x 26 |
| Shortcut Layer | | | |
| Convolution | 512 | 3 x 3/ 2 | 13 x 13 |
| Convolution | 1024 | 3 x 3 | 13 x 13 |
| Max | | 5 x 5 | 13 x 13 |
| Route | | | 13 x 13 |
| Max | | 9 x 9 | 13 x 13 |
| Route | | | 13 x 13 |
| Max | | 13 x 13 | 13 x 13 |
| Route | | | 13 x 13 |
| Convolution | 256 | 1 x 1 | 13 x 13 |
| Convolution | 512 | 3 x 3 | 13 x 13 |
| Convolution | 21 | 1 x 1 | 13 x 13 |

| YOLO | | | |
|---|---|---|---|
| Route | 23 | | 13 x 13 |
| Convolution | 128 | 1 x 1 | 13 x 13 |
| Upsample | | 2x | 26 x 26 |
| Route | | | 26 x 26 |
| Convolution | 256 | 3 x 3 | 26 x 26 |
| Convolution | 21 | 1 x 1 | 26 x 26 |
| YOLO | | | |

*3.5 Loss Functions for Bounding Box Regression*

The loss functions used in the networks are Generalized IoU (GIoU) and Mean square error (MSE) loss. The bounding box regression is an essential task in vehicle detection. Vehicle detection and localization usually depend on accurate bounding box regression. IoU loss function is the commonly used technique in object detection algorithms, but it has an issue with non-overlapping objects. GIoU for bounding box regression addresses this issue. Hence, it is an optimal choice to use GIoU into the network during the training process.

Let us consider $B^p = (x^p_1, y^p_1, x^p_2, y^p_2)$ and $B^g = (x^g_1, y^g_1, x^g_2, y^g_2)$ are coordinates of predicted bounding boxes and ground truth bounding boxes. The GIoU loss function for these coordinates is derived in the equations given below from [29].

$$IoU = \frac{I}{U}, \text{ where } U = A^p + A^g - I \qquad (1)$$

$$GIoU = IoU - \frac{A^c - U}{A^c} \qquad (2)$$

$$\int_{IoU} = 1 - GIoU \qquad (3)$$

GIoU loss is useful when the objects are not overlapped with each other. It is invariant to scale the same as of IoU loss function. The gradient of IoU loss is zero for the non-overlapping objects, whereas GIoU forms gradients in all objects even if the objects are non-overlapping.

Mean square error loss function is commonly used for regression in most of the networks. MSE is the sum of squared distances between the ground truth object and predicted object values. Its general equation is shown below. In the equation, $y_i$ is the ground truth object value and $y^p_i$ is the predicted object value and n is the number of classes.

$$MSE = \frac{\sum_{i=1}^{n}(y_i - y_i^p)^2}{n} \quad (4)$$

**Table 3** Train and Test image Datasets

| Dataset | Number of Classes | Classes | Train Images | Test Images |
|---|---|---|---|---|
| PASCAL VOC 2007, 2012 | 2 | Car and Bus | 16,551 | 4952 |
| MS COCO 2014 | 3 | Car, Bus and Truck | 11,432 | 5545 |
| PASCAL VOC + MS COCO | 3 | Car, Bus and Truck | 27,983 | 10,497 |

*3.6 Datasets:*

The datasets used in this work are PASCAL VOC 2007, PASCAL VOC2012, MS COCO 2014 [11,12] for training and GRAM Road-Traffic Monitoring (GRAM-RTM) video dataset for detection result analysis [32]. PASCAL VOC contains a total of 20 different classes, and MS COCO has 80 classes, these datasets are usually used for object detection models for training with all classes present in the datasets. Training the vehicle detection network with all the classes in the dataset is not an ideal choice for building a good vehicle detector. Hence, a combined dataset of car, bus classes from PASCAL VOC and car, bus, truck classes from MS COCO is used in training. The combined dataset of vehicle detection contains a total of 27,983 training images and 10,497 testing images. In Table 3, all the classes and the number of images used in the study are tabulated.

The evaluation of these models is based on the metrics of Mean of average precision mAP with a value of 0.5, Network weight size, Input size, BFLOPs, and inference time based on FPS. The models also compared with two different input sizes 416×416 and 640×640, which are processed on Tesla P100-PCIE-16GB GPU. The Experimental platform configuration is listed in Table 4.

**Table 4** Experimental Platform Configuration

| Computing Machine | Configuration |
|---|---|
| Operating System | Ubuntu 18.04.3 LTS |
| GPU | Tesla P100-PCIE-16GB |
| RAM | 16 |
| GPU acceleration library | CUDA10.0, CUDNN7.4 |

*3.7 Network Training and Experimental Analysis*

The training process of LittleYOLO-SPP is carried out by various methods with three different datasets. In the first training method, LittleYOLO-SPP is trained on PASCAL VOC 2007, 2012 datasets using SGD. The unrequired classes from the datasets are removed, then vehicle-based classes car and bus are used. The input size of the network is 416 × 416. The batch size 64 and mini-batch size 8 is used. The total train images in the dataset are 16,551 and the test set of images is 4952. The momentum is 0.9; the learning rate is 0.001 and weight decay of 0.0005 is set as default parameter in this model. The number of iterations processed by the network is close to 150 thousand. The hyperparameters used on the YOLO detection layers are mse loss, scaling, and IOU thresholding of 0.50. Multiscale training is used to train the images with different sizes regardless of the original size. This training method achieves mAP of 77.44% higher than YOLOv2-tiny and YOLOv3-tiny, as shown in Table 5. The network also achieves the best inference time of about 5ms for 177 FPS.

**Table 5** LittleYOLO-SPP Performance results on PASCAL VOC 2007, 2012 Dataset

| Network | AP (car) | AP (bus) | mAP |
|---|---|---|---|
| YOLOv2-tiny | 68.15 | 68.85 | 68.5 |
| YOLOv3-tiny | 77.50 | 73.0 | 75.25 |
| **LittleYOLO-SPP (Proposed)** | 79.31 | 75.57 | **77.44** |

In the second training model, MS COCO 2017 dataset with car, bus, and truck classes was used. The input size and all other parameters and functions used are the same as the previous training method. Table 6 summarizes the comparison results of the five models. These five models are trained on same configuration settings of LittleYOLO-SPP. Experimental results shows that LittleYOLO-SPP achieves much better results in average test time. It also achieves the result of 52.95% mAP, which is better than existing network models as can be seen in Table 6. The LittleYOLO-SPP improved the detection accuracy by 6.83% compared with YOLOv3-tiny.

**Table 6** LittleYOLO-SPP Performance results with other Networks on MS COCO 2014 Dataset.

| Network | Number of Classes | FPS | mAP |
| --- | --- | --- | --- |
| Faster-RCNN | 3 | 198 | 34.89 |
| EfficientNet Lite | 3 | 85 | 43.95 |
| Tiny YOLO | 3 | 244 | 31.44 |
| YOLOv3-tiny | 3 | 220 | 46.12 |
| **LittleYOLO-SPP (proposed)** | 3 | 49 | **52.95** |

In the final training method, the datasets of PASCAL VOC and MS COCO are combined. A total of 27983 training images were selected from the three different classes such as car, bus, and truck. The network is trained with an input size of 416 × 416 and 640 ×640. In 416 × 416 models, the batch size is 64, and mini-batch is 8, the learning rate of 0.001, and weight decay 0.0005.MSE loss and GIoU loss functions are used separately while training the networks. In 640 × 640 models, the batch size is increased to 90 and mini-batch to 32. Then followed all other default settings from the last training network and used GIoU loss. The iterations processed by the network is between 100 to 150 thousand iterations. The network uses various hyperparameters in YOLO detection layers are GIoU loss, scaling factor for bounding boxes 1.05, IoU threshold of 0.223, IoU normalization 0.07, and class normalization 1.0.

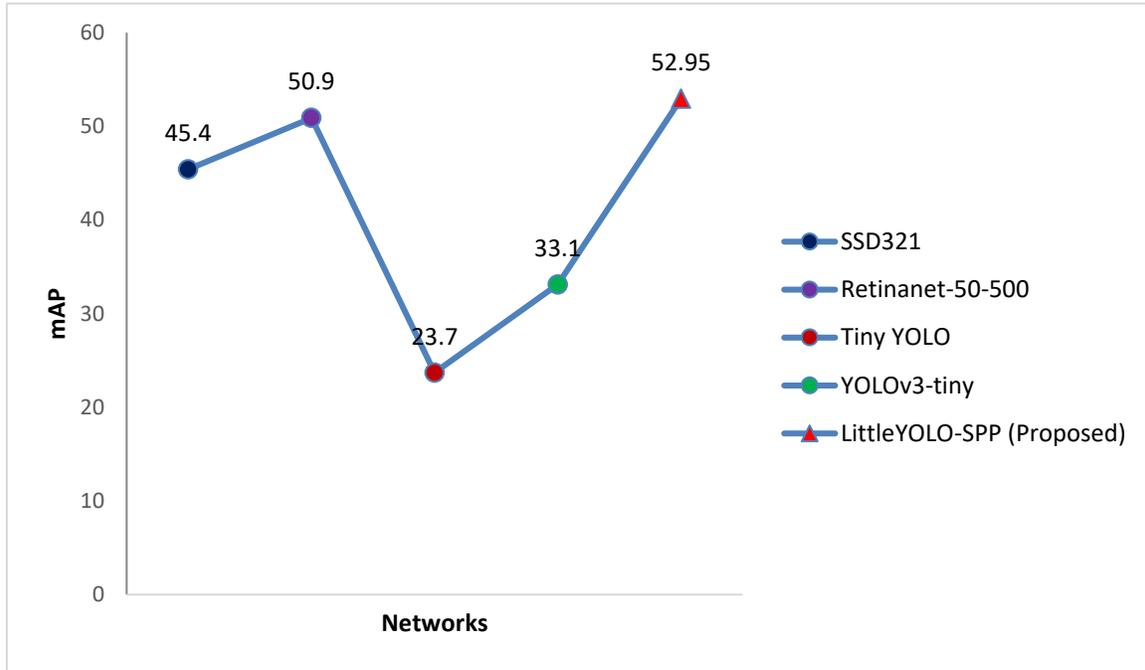

**Fig. 2.** Comparison graph for LittleYOLO-SPP with other network models on MS COCO 2014 dataset.

The batch normalization on all convolutional layers is kept as it is from YOLO to optimize the network, which is also proven to be effective in improving mAP [14]. The Mish activation function is used rather than using leaky relu or swish functions [33]. LittleYOLO-SPP-640 achieves high mAP of 47.41% than the other two models, which records 46.05% and 38.43% shown in Table 7, and its corresponding graph is drawn in Fig. 2. The combined dataset network achieves less mAP than PASCAL VOC and MS COCO datasets.

**Table 7** LittleYOLO-SPP Performance results on PASCAL VOC 2007, 2012 and MS COCO 2014 Combined Dataset

| Network | Loss | AP (car) | AP (bus) | AP (Truck) | FPS | mAP |
|---|---|---|---|---|---|---|
| **LittleYOLO-SPP-416 (proposed)** | MSE | 30.60 | 54.78 | 29.90 | 50 | 38.43 |
| **LittleYOLO-SPP-416 (proposed)** | GIoU | 42.47 | 60.27 | 35.41 | 46 | 46.05 |
| **LittleYOLO-SPP-640 (proposed)** | GIoU | 48.99 | 57.61 | 35.64 | 25 | **47.41** |

LittleYOLO-SPP network has a model size of 49.77 MB, and the Computational power required by the model is 16.128 BFLOPS/s. This model, trained with PASCAL achieved a high inference time of 5 seconds.

Then other models are performed with average inference time of 20 seconds. LittleYOLO-SPP-640 records 25 FPS in combined datasets. The detection of the network is tested in online live road traffic video feed and also tested the performance on GRAM-RTM road dataset.

The LittleYOLO-SPP with input size of 416 × 416 trained with PASCAL VOC 2017, 2012 dataset prediction results are shown in Fig. 3. From the predicted CCTV video frame, the network is analyzed, and it is capable of detecting long-range small size noise vehicles quickly. In Fig. 6, LittleYOLO-SPP-416 trained with MS COCO 2014 dataset shows that the network can predict many nested vehicles accurately without missing a single car. The network of LittleYOLO-SPP is simple and less complex than the original YOLOv3-tiny network.

Error analysis of YOLOv3-tiny compared with LittleYOLO-SPP shows that YOLOv3-tiny makes many errors based on close localization. It also struggles to detect vehicles in long distances. Thus LittleYOLO-SPP focuses mainly on improving these issues. In PASCAL VOC dataset LittleYOLO-SPP makes some errors on small vehicles, then with MS COCO errors are mainly with medium, large vehicles, and also background error occurs. These errors are mostly rectified while combining two datasets but mAP dropped when compared with MS COCO. The networks architecture also altered by replacing max-pooling layers with residual connections which performs better than the baseline network. The SPP Block, MSE loss, GIoU loss, Batch Normalization and Non-maximum suppression are used for optimizing the network. While evaluating these networks trained with PASCAL VOC, MS COCO and Combined Datasets, the network trained on MS COCO dataset provides better results in terms of accuracy, and combined dataset performance is high.

This work shows that the model is specifically designed for the vehicle detection task. Many object detection networks are not focused entirely on vehicles. In this work, other than vehicle classes are deleted from the dataset. Vehicle classes such as a car, bus, and truck are used to train this network. Hence training speed and learning of vehicle-based features are quite good in this method. Many YOLO based object detection networks have high computational cost and its training speed also slow due to its complex and its deep layers. In YOLOv3-tiny the feature extraction convolutional layers are separated by max-pooling by removing those max pooling and adding a spatial pyramid pooling boosts the performances of the network.

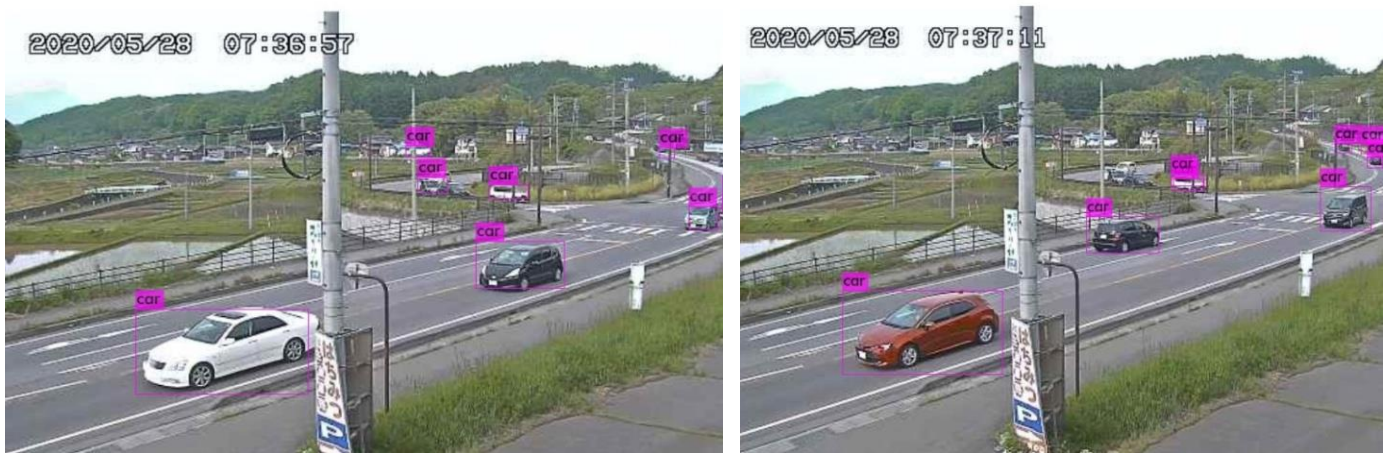

**Fig. 3.** Detection results of LittleYOLO-SPP trained on PASCAL VOC 2007, 2012 dataset

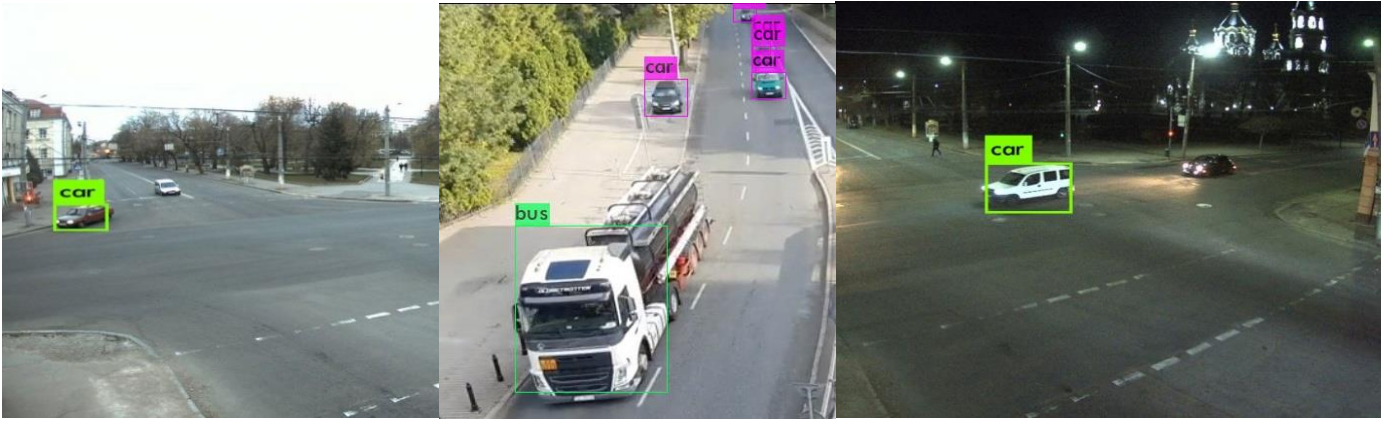

**Fig. 4.** Detection errors of LittleYOLO-SPP

The network also incorporated two loss functions MSE which are commonly used and GIoU best loss function for detection tasks. GIoU further increases the accuracy of the network. K means ++ clustering algorithm is used to define the anchor boxes, which is essential in the detection part.

The network is tested on both online and offline modes. The inference time of the proposed network is high, on live CCTV frames which makes it suitable for real-time predictions. The network training time also less when compared with other networks. The YOLOv3-tiny evaluated with 400 thousand iterations which are approximately double when compared with LittleYOLO-SPP because it achieved its maximum mAPwhile, reaching less than 150-200 thousand iterations.

In Fig. 4 shows the network missed to predict some vehicles and also it mistakenly captioned a gas tanker as a bus. The front end of the tanker has features of the bus but backside the tanker features not fed into training. The dataset PASCAL and MS COCO doesn't contain images of tanker; hence the network couldn't detect it properly. The missed vehicles in the frames are due to fewer images in the dataset. To increase the detection accuracy of the network, more vehicle-based images are needed during training.

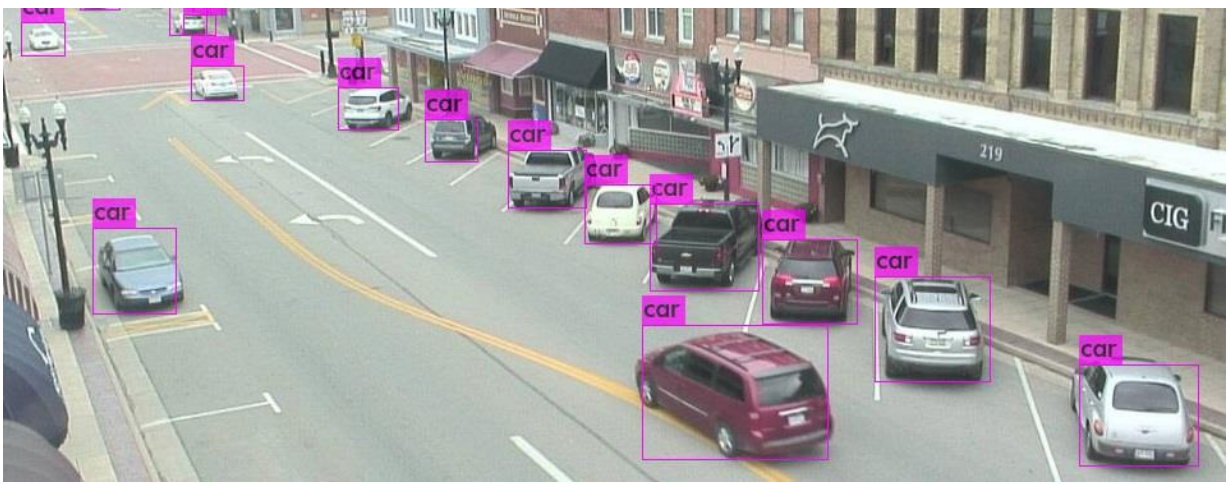

**Fig. 5.** Detection results of LittleYOLO-SPP trained on MS COCO 2014 dataset

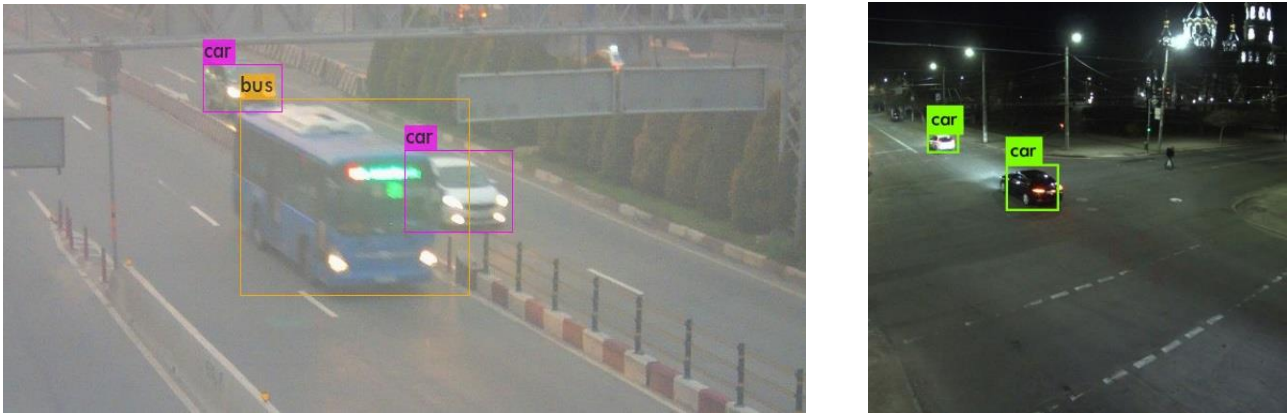

**Fig. 6.** Detection results of LittleYOLO-SPP-416 trained on PASCAL VOC 2007, 2012 and MS COCO 2014 dataset

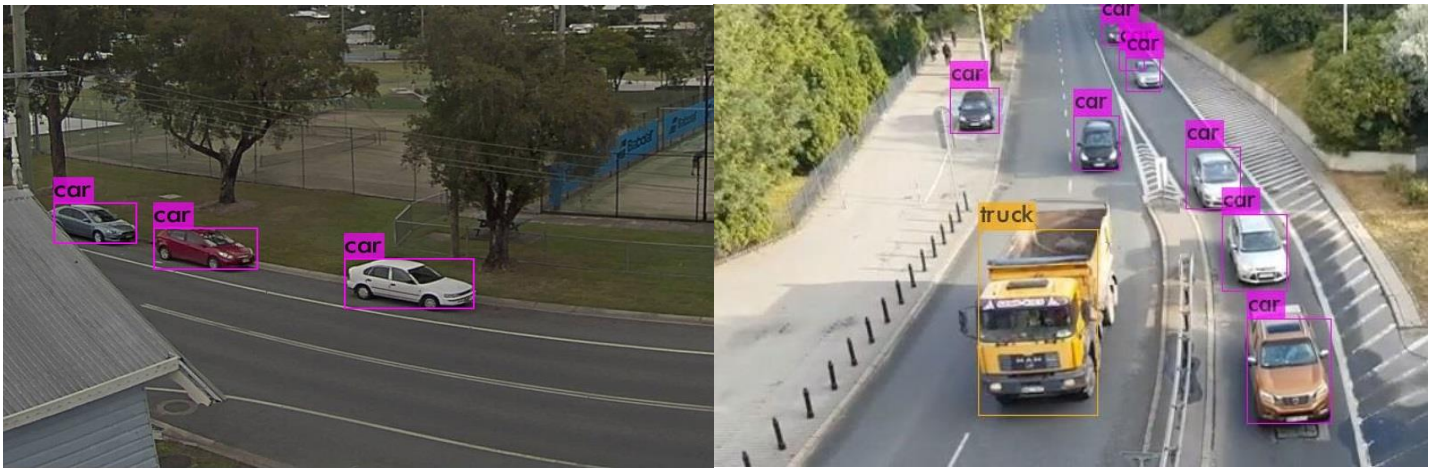

**Fig. 7.** Detection results of LittleYOLO-SPP-640 trained on PASCAL VOC 2007, 2012 and MS COCO 2014 dataset

From Fig.5, 6 and Fig. 7, the network detection results are quite good. In Fig. 6, the weather condition is terrible, and video feed clarity also not good, but the network predicts the vehicles with high accuracy. In Fig. 3, the network predicts small vehicles very far away from the CCTV camera. Hence the network is very effective in the detection of vehicles.

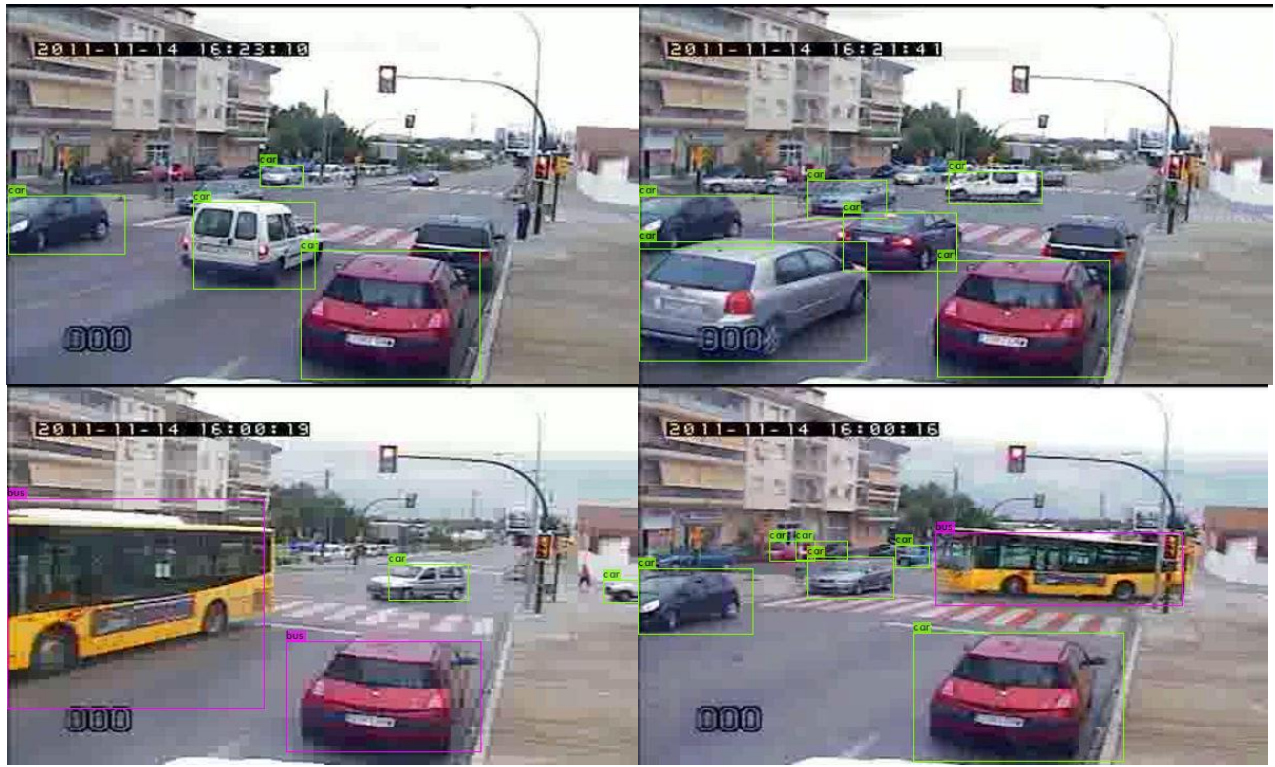

**Fig. 8.** Detection results of LittleYOLO-SPP models trained on PASCAL VOC 2007, 2012, and MS COCO 2014 dataset, which is tested with GRAM-RTM dataset.

In Fig. 6 the CCTV video feed is very noisy, two vehicles are occluded with one other, and the weather condition is low light. Still, the proposed model can easily predict the vehicles in the video frames as bus and car. The detection results of GRAM-RTM dataset on Urban CCTV video feed frames are shown in Fig. 8. By experimenting and analyzing all these results, the proposed vehicle detector network LittleYOLO-SPP is robust and very fast in detecting vehicles accurately.

There are many variants of YOLO networks available, but they focus on multiple object detection, pedestrian detection, aerial view detection, etc. They also lack performance in complex environments and at night time. In the proposed model, the network can predict vehicles in darker environments, but Yolo-v3 tiny cannot predict vehicles or objects in darker environments. This illustrates that the improved network has good adaptability to the complex real-time environment of the road. The network also missed to detect some small vehicles, but its accuracy is much improved than the base network of Yolov3-tiny.

**4. Conclusion**

In this paper, LittleYOLO-SPP an efficient algorithm for detecting vehicles based on the YOLOv3-tiny object detection network is proposed. The modified LittleYOLO-SPP consists of improved feature extraction layers, spatial pyramid pooling, and GIoU loss function. K-means++ clustering algorithm is used to create the anchor box sizes. The proposed network achieves the mAP of 77.44% in PASCAL VOC 2007,2012 and 52.95% in the MS COCO dataset. The experimental results show that the network is speedy, achieving an inference time of 5ms. LittleYOLO-SPP detects vehicles in real-time from traffic CCTV camera feeds with high accuracy, which makes the network suitable for high-end surveillance systems. However, the algorithm is still missed to detect some vehicles and made few labelling mistakes too. The goal of future research includes addressing these issues with more vehicle classes.